\pdfoutput=1
\documentclass[11pt]{article}

\usepackage{acl}

\usepackage{times}
\usepackage{latexsym}

\usepackage[T1]{fontenc}
\usepackage[utf8]{inputenc}

\usepackage{microtype}

\usepackage{graphicx}
\usepackage{caption}
\usepackage{subcaption}

\usepackage{amssymb}

\usepackage{url}

\title{Archive TimeLine Summarization (ATLS): Conceptual Framework for Timeline Generation over Historical Document Collections}

\author{Nicolas Gutehrlé \\
    Laboratoire CRIT, \\
    University of Bourgogne 
    \\ Franche-Comté, France\\ %30 rue Mégevand, 25000, Besançon, France\\
    \texttt{nicolas.gutehrle} \\  \texttt{@univ-fcomte.fr} 
    \And
    Antoine Doucet \\
    Laboratoire L3i, \\ University of La Rochelle, \\
    France\\
    \texttt{antoine.doucet} \\
    \texttt{@univ-lr.fr} \\
    \And
    Adam Jatowt \\
    Dept. of Computer Science \& \\
    Digital Science Center,\\
    University of Innsbruck, Austria \\
    \texttt{adam.jatowt} \\
    \texttt{@uibk.ac.at} \\
  }
  
\begin{document}
\maketitle

\begin{abstract}

   Archive collections are nowadays mostly available through search engines interfaces, which allow a user to retrieve documents by issuing queries. The study of these collections may be, however, impaired by some aspects of search engines, such as the overwhelming number of documents returned or the lack of contextual knowledge provided. New methods that could work independently or in combination with search engines are then required to access these collections. In this position paper, we propose to extend \textit{TimeLine Summarization} (TLS) methods on archive collections to assist in their studies. We provide an overview of existing TLS methods and we describe a conceptual framework for an \textit{Archive TimeLine Summarization} (ATLS) system, which aims to generate informative, readable and interpretable timelines. 
%   We make several suggestions to implement this framework and describe potential use cases.
   
\end{abstract}

% =================================================================
\vspace{-.5em}
\section{Introduction}
\label{sec:introduction}
\subsection{Exploring archives}

In the recent years, archives and libraries across the world have frequently conducted digitization campaigns of their collections. This first opened access to thousands of historical documents to a wider public, but also propelled the emergence of new research fields such as Digital Humanities and Digital History. These collections are usually accessible through search engines, which return documents relevant to a query specified by the user. Unfortunately, standard search engines are not fully suited to assist users in exploring historical collections such as news archives where temporal aspects of documents play a key role. Firstly, search engines return documents by their relevance to the query, typically without considering the chronological or causal relations between them, which may prevent the user from understanding the interrelations between events. Furthermore when exploring such documents, the user might lack the contextual knowledge to understand the events that are mentioned in them. This is especially true when exploring news archives coming from distant pasts or exploring longitudinal collections, i.e. which span over a long time frame such as decades or centuries. Search engines do not seem to consider the importance of an event mention for a given query, thus less important events might be returned by the system, especially for broad queries. Improved search engines are then required to study such collections.

\vspace{-.5em}
\subsection{Augmenting search engines with timelines}
\begin{figure*}
    \centering
    
    \begin{subfigure}[b]{.5\textwidth}
        \centering 
        \includegraphics[width=\textwidth]{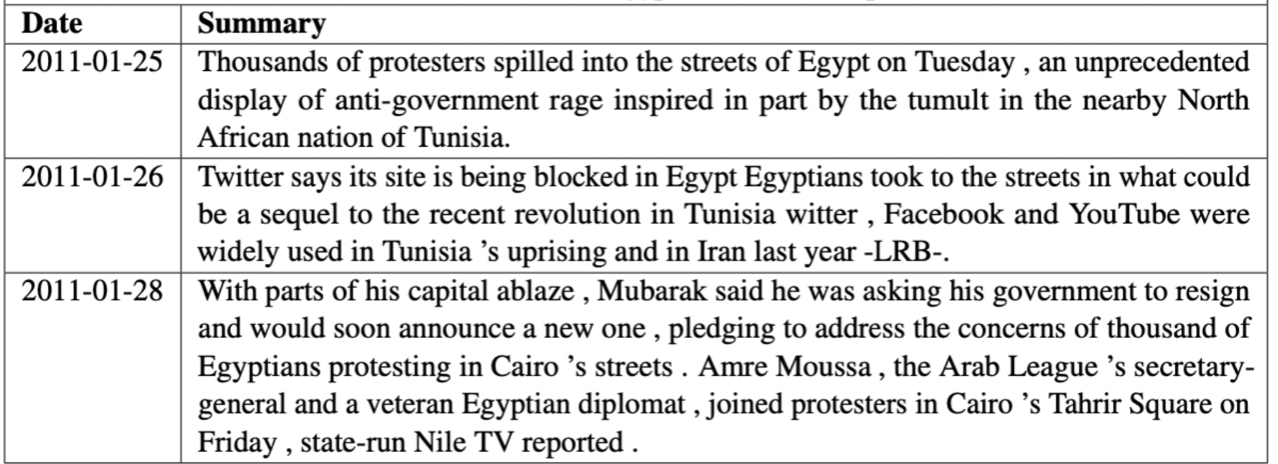}
    \end{subfigure}%
    \begin{subfigure}[b]{.5\textwidth}
        \centering 
        \includegraphics[width=.9\textwidth]{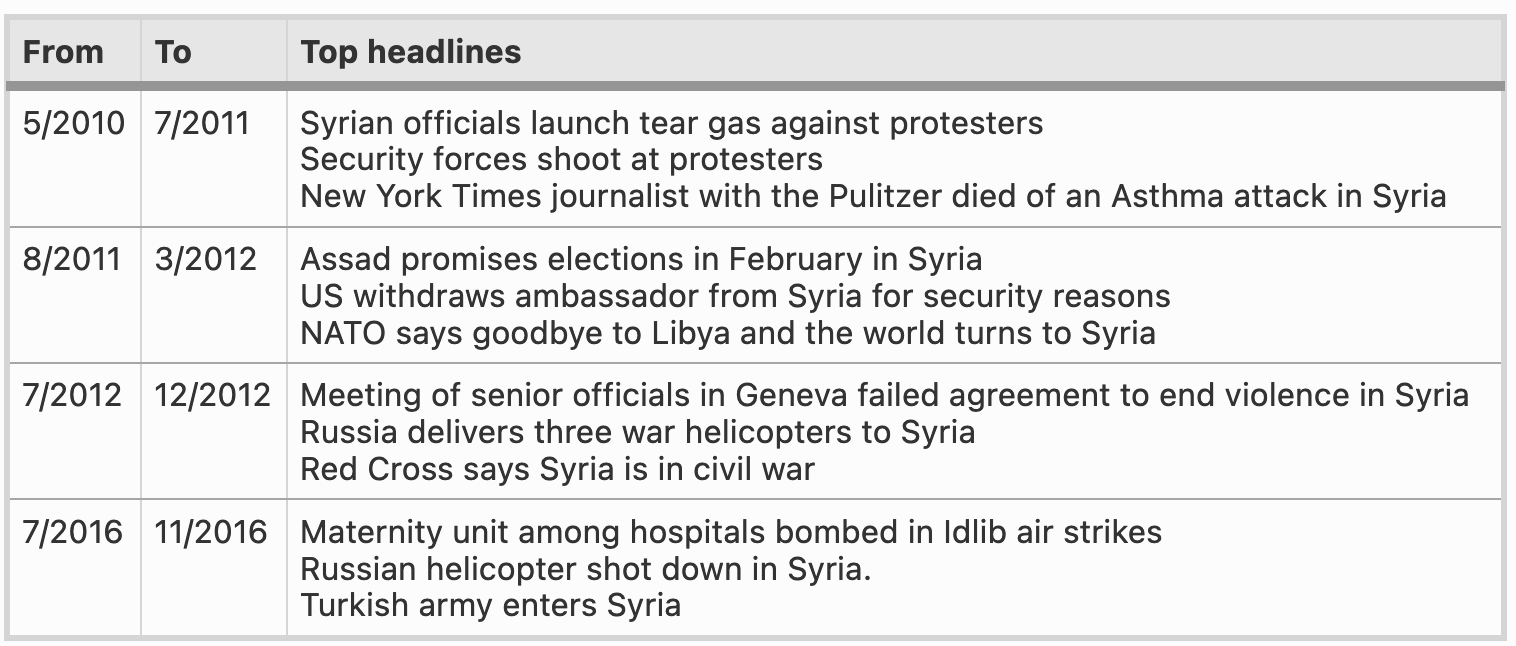}
    \end{subfigure}%
    
    \caption{Examples of generated timelines by \citet{yu-etal-2021-multi} (left) and \citet{10.1007/978-3-319-76941-7_80} (right), summarizing a set of documents about respectively Egyptian protests and the Syrian War. The left timeline outputs a summary on a day-to-day basis, whereas the right timeline lists events using uneven periods of time.}
    \label{fig:exampletl}
\end{figure*}

\vspace{-.5em}
One promising method to improve the output of search engines operating over archival collection is \textit{TimeLine Summarization} (TLS). TLS consists in summarizing multiple documents by generating a timeline where important events detected in the dataset are associated with a time unit such as a day. TLS is a subfield of the \textit{Multi-Document Summarization} (MDS) task and has been studied extensively in the NLP community: for instance, \citet{swan-allan-overwiew-2000} generate clusters of Named Entities and noun chunks that best describe major news topics covered in a subset of the TDT-2 dataset \citep{10.1145/290941.290954}, which contains text transcripts of broadcast news spanning from January 1, 1998, to June 30, 1998, in English; \citet{nguyen-etal-2014-ranking} generate timelines by detecting events that are the most relevant to a user query. They apply their methodology on a dataset of newswire texts in English covering the 2004-2011 period provided by the AFP French news agency; \citet{duan-jatowt-katsumi-multitl-2017} extend these methods to summarize the common history of similar entities such as Japanese Cities or French scientists. Examples of timelines generated by such methods are shown in Figure \ref{fig:exampletl}.

Hence, TLS could serve as a distant reading tool and as a first step in exploring a dataset by providing an overview of its key events. Moreover, TLS could be combined with search engines and used as an interface to search results returned by issuing queries over large datasets, as suggested in \citet{swan-allan-overwiew-2000, DBLP:conf/desires/2021}. From there, the user could zoom into the documents in order to proceed to close reading. Furthermore, these summaries would be presented in chronological order, thus preserving the link between events, and could also be contextualized by adding data from external knowledge bases as in \citet{10.1145/2600428.2609526}.

Search engines augmented with timelines would be especially useful in a Digital Humanities (DH) context such as for facilitating the study of historical datasets, as they would provide necessary context to understand past events and to structure the event landscape. They could also help the user understand the history of a particular entity such as a person or a location, or even a group of such entities through providing a bird's-eye view of the relevant data. A good example of such search engine augmented with TLS is the Conta-me Histórias (Tell me stories) platform\footnote{\url{https://contamehistorias.pt/arquivopt}}, where the user can query news articles from the Portuguese web archive. The user-friendly interface allows a distant reading of the documents returned by the query through a timeline that summarizes them, but also allows close reading by preserving the link to the original documents. To the best of our knowledge, works on applying
%no previous work has 
%applied 
TLS methods to structure archives of historical documents, or more broadly in the Digital Humanities field, are quite scarce.
% nor has applied it in the Digital Humanities and Digital History domains.

\subsection{Challenges of applying TLS to archives}

Unfortunately, several aspects of such archives make the application of TLS methods not straightforward: first, these datasets are often processed with Optical Character Recognition (OCR). Previous studies have shown that downstream tasks such as Named Entity Recognition (NER), Event Detection (ED) \citep{boros:hal-03635985}, Topic Modelling (TM) \citep{mutuvi:hal-03025563} or Named Entity Linking (NEL) \citep{linharespontes:hal-02557116} are impacted by the quality of the OCR output. To our knowledge, there is no study on the impact of OCR on TLS, but we can assume it will be similar. Furthermore, archive collections may also differ from contemporary data because of their temporal context: orthographic rules may differ, places might have changed names \citep{10.5555/646634.699911} or concepts may have acquired another meaning. Most existing annotated resources necessary for NLP components such as NER or ED are created on contemporary data. Historical documents archives are thus harder to process because of this lack of suitable annotated resources.

Most TLS methods generate timelines through statistical analysis of the input dataset. They also often require that the input corpus contains documents of a similar type and similar content. However, an archive collection may be heterogeneous and contain documents of different authors, genres, topics and periods. It may also be fragmentary and not as complete as a contemporary dataset. Finally, although the timelines generated by TLS systems are often easy to read, the process that created them is often not made explicit. If timelines must assist the study of historical datasets by highlighting important events, they must be interpretable and explain why these events are deemed important.

In this position paper, we propose to extend \textit{TimeLine Summarization} (TLS) methods to assist in the studies of archive collections. We first present an overview of existing TLS methods. We then describe a conceptual framework for an \textit{Archive TimeLine Summarization} (ATLS) system, which aims to generate informative, readable and interpretable timelines, before suggesting several methods to implement it.

% . We make several propositions to implement it in an unsupervised fashion, so as to account for the lack of suitable annotated resources. 

This paper is organized as follows: in Section \ref{sec:relatedwork} we present an overview of existing TimeLine Summarization methods. In Section \ref{sec:methodology} and \ref{sec:discussion}, we respectively describe our conceptual framework and discuss some of its potential applications. Finally, we present our conclusion in Section \ref{sec:conclusion}, alongside possibilities for future works.

% =================================================================
\vspace{-.5em}
\section{Related Work}
\label{sec:relatedwork}

\vspace{-.5em}
\subsection{TimeLine Summarization}

Most TLS methods generate timelines by applying the two following steps: the \textbf{Date Selection} step which identifies and ranks the key dates in the documents, and the \textbf{Date Summarization} step which generates a summary of an event occurring at a specific date by picking important sentences in the documents published on that date. To identify important dates in the dataset, \citet{gholipour-ghalandari-ifrim-2020-examining} select the most frequent date mentions, \citet{tran-etal-2015-joint} use a graph-ranking model and \citet{Kessler2012FindingSD} combine a clustering model and a supervised classifier. For the second step, \citet{quatra-al-datefirst-2021} apply state-of-the-art methods for Text Summarization (TS) such as TextRank \cite{mihalcea-tarau-2004-textrank} whereas \citet{martschat-markert-2018-temporally} adapt methods from the Multi-Document Summarization (MDS) field. TLS has been generally extractive, i.e. the summary is created by copying textual elements (e.g., sentences or paragraphs) from the input data \citep{tran-alrifai-herder-headlines-2015}. Other works are abstractive, i.e. the summary is a completely new text generated by the system \citep{steen-markert-2019-abstractive}. 

TLS methods in general tend to be applied to summarize datasets describing large events, such as the Egyptian protests or the Syrian War \citep{tran-etal-2015-joint, martschat-markert-2018-temporally}. These methods require that the dataset covers a constrained period of time and is homogeneous, i.e. that the documents cover the same topic. Standard TLS methods are thus not suited to summarize heterogeneous or longitudinal datasets. Some works such as \citet{nguyen-etal-2014-ranking, Kessler2012FindingSD, 10.1145/1008992.1009065, 10.1007/978-3-030-15719-7_34} can be described as \textit{Query-based TimeLine Summarization} (QTLS), as they apply TLS on documents related to a user query such as documents returned by a search engine. 

QTLS generally consists in the two following steps: \textit{Event Detection} and \textit{Event Ranking}. To detect events, \citet{10.1145/1008992.1009065} select any sentence where the terms of the query appear, \citet{nguyen-etal-2014-ranking} cluster by a common date every sentence returned by the query and \citet{10.1007/978-3-030-15719-7_34} detect peaks of date occurrences in the time span covered by the documents. Other works train a classifier to detect important events \citep{chasin-2010-event-wiki-linea} or rank events by their importance with a Learning-to-Rank model \citep{ge-etal-2015-bring}. However, these classifiers need training data, which are difficult to create since defining what is important is a subjective matter. This can lead to disappointing results as shown in \citet{chasin-2010-event-wiki-linea}.
To determine the importance of events, \citet{nguyen-etal-2014-ranking} first score them according to their relevancy and saliency to the query, then rerank them to ensure a diverse timeline. \citet{10.1145/1008992.1009065} rank the importance of a sentence according to their "interest" and "burstiness", then remove duplicate sentences to ensure diversity. \citet{10.1007/978-3-030-15719-7_34} use the keyword extractor YAKE! \citep{10.1007/978-3-319-76941-7_80} to weight the terms in the event description. Duplicate event descriptions are detected with the Levenshtein similarity measure and removed. Those methods finally select the top most important events to generate the timeline.

In order to generalize the application of TLS, \citet{yu-etal-2021-multi} propose a Multiple TimeLine Summarization (MTLS) system, which generates a timeline for each story found in the dataset. To do so, it first detects events mentioned in the dataset and measures their saliency and consistency. An event linking step determines the link between these events in order to generate each timeline. Similarly, \citet{Duan2020ComparativeTS} propose the \textit{Comparative TimeLine Summarization} (CTLS) task, which generates a comparative timeline highlighting the contrast between two timestamped timeline documents (e.g. biographies, historical sections, ...) by computing local and global importance of events.
% To do so, they propose the D-APMRRW model, which computes the importance of events locally and globally in the covered period. 

There are few datasets for the TLS task such as 17 Timelines (T17) \citep{Tran2013LeveragingLT}, CRISIS \citep{tran-alrifai-herder-headlines-2015}, ENTITIES \citep{gholipour-ghalandari-ifrim-2020-examining}, CovidTLS  \citep{quatra-al-datefirst-2021} or TLS-Covid19 \citep{Pasquali2021TLSCovid19AN} which are constructed from contemporary news articles. However, datasets are often lacking in most projects. It is then necessary to create a dataset from scratch as in \citet{minard-etal-2015-semeval, nguyen-etal-2014-ranking, ge-etal-2015-bring, bedi-etal-2017-event} or extend existing ones as in \citet{yu-etal-2021-multi}.
% in order to experiment on the common settings of DH. 

Due to this lack of datasets, evaluating TLS systems is a difficult task. The date selection step can be evaluated with the F1-measure \citep{quatra-al-datefirst-2021, gholipour-ghalandari-ifrim-2020-examining} or with the Mean Average Precision (MAP) metric \citep{nguyen-etal-2014-ranking}. The date summary is often evaluated with one of the ROUGE metrics \citep{lin-2004-rouge} to compare a ground-truth timeline and a generated one \citep{nguyen-etal-2014-ranking, Duan2020ComparativeTS, yu-etal-2021-multi, gholipour-ghalandari-ifrim-2020-examining}. Methods relying on event detection such as \citet{ge-etal-2015-bring, minard-etal-2015-semeval, bedi-etal-2017-event} often evaluate their system in terms of Precision, Recall and F1-measure. However, most projects often lack datasets and must then resort to human evaluation as in \citet{duan-jatowt-katsumi-multitl-2017, swan-allan-overwiew-2000, tran-alrifai-herder-headlines-2015}.

\subsection{TLS Variants}

We present below formal definitions of several existing TLS variants: 
\vspace{-.5em}
\begin{description}
    \item[TLS:] takes as input a standalone homogeneous dataset of timestamped documents $\mathcal{D} = \{d_1, d_2, ..., d_{|D|}\}$ and generates a timeline $T = \{p_1, p_2, ..., p_{|T|}\}$ of time-summary pairs $p_i = (t_i, s_i)$, where $s_i$ summarizes important events happening at time $t_i$;
    \vspace{-.5em}
    
    \item[QTLS:] outputs a timeline $T = \{p_1, p_2, ..., p_{|T|}\}$ as a sequence of time-summary pairs $p_i = (t_i, s_i)$ from a set of timestamped documents $\{d_1, d_2, ..., d_{|D|}\}$ based on a query $\mathcal{Q} = \{w_1, w_2, ..., w_k\}$ where $w_i$ denotes a word belonging to the query;
    \vspace{-.5em}
    
    \item[MTLS:] takes as input a dataset of timestamped documents $\mathcal{D} = \{d_1, d_2, ..., d_{|D|}\}$ that can be standalone or returned using a query $\mathcal{Q} = \{w_1, w_2, ..., w_k\}$, and outputs a set of timelines $\mathcal{T} = \{T_1, T_2, ..., T_m\}$ for each story or topic detected in $\mathcal{D}$, where each timeline $T_i$ is a sequence of time-summary pairs $p_i = (t_i, s_i)$;
\vspace{-.5em}

    \item[CTLS:] takes as input two datasets of timestamped documents $\mathcal{D_A} = \{d_1, d_2, ..., d_{|D_A|}\}$  and $\mathcal{D_B} = \{d_1, d_2, ..., d_{|D_B|}\}$ and outputs two timelines $\mathcal{T_A}$ and $\mathcal{T_B}$ made of contrasting events detected in $\mathcal{D_A}$ and $\mathcal{D_B}$, each as a sequence of time-summary pairs $p_i = (t_i, s_i)$;
\vspace{-1em}
    
\end{description}

% =================================================================

\section{Framework}
\label{sec:methodology}
\vspace{-.5em}
In this section, we present a conceptual framework for an \textit{Archive TimeLine Summarization} (ATLS) which addresses the challenges raised by archive collections such as the sparsity of data, OCR problems, context shifts and linguistic changes over time in order to generate timelines based on these datasets. We first provide a definition of ATLS and describe the type of dataset expected before presenting the framework and discussing how to evaluate its output.

\vspace{-.5em}
\subsection{Overview}
\label{sec:overview}

\begin{figure*}
    \centering
    \includegraphics[width=.9\textwidth]{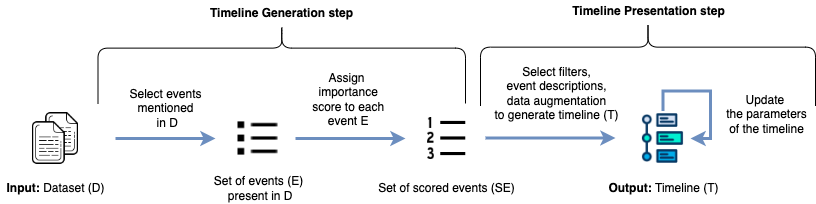}
    \caption{Conceptual pipeline for building the ATLS system}
    \label{fig:tlgeneration}
\end{figure*}

The framework consists of the two key steps: \textit{Timeline Generation} and \textit{Timeline Presentation}. The first step extracts textual elements describing an event and attributes them an importance score. The second one generates the timeline by filtering events and selecting their description. 
% Although this overall approach is similar to the QTLS systems presented in Section \ref{sec:relatedwork}, it can be applied to a dataset returned by a query or a standalone dataset provided by the user.

The processing stages of the framework are shown in Figure \ref{fig:tlgeneration}. The first step has to run only once over the processed dataset, since it aims to detect the elements composing the timeline to be generated. In contrast, the second step can be run multiple times to update the timeline.

\vspace{-.5em}
\subsection{Problem Definition}

We define ATLS as follows:

\vspace{-.5em}
\begin{description}
    \item[Input:] A longitudinal dataset of timestamped documents $\mathcal{D} = \{d_1, d_2, ..., d_{|D|}\}$ taken from an archival collection, either standalone or returned by a query $\mathcal{Q} = \{w_1, w_2, ..., w_k\}$. The period of time covered by $\mathcal{D}$ is usually much longer than the one typically used in TLS.
    
    \item[Output:] A timeline $\mathcal{T}$ generated from $\mathcal{D}$ as a sequence of time-summary pairs $p_i = (t_i, s_i)$, where $s_i$ summarizes important events happening at time $t_i$.
    % The length $|\mathcal{T}|$ of the timeline is usually longer than the one common for TLS.

\end{description}
 \vspace{-.5em}
 We compare the key characteristics of TLS and ATLS in Tab. \ref{tab:comparisontable}. 
% ===========================================================
\vspace{-.5em}
\subsection{Expected Dataset}
\label{sec:dataset}

The framework takes as input a longitudinal dataset composed of timestamped documents, such as news articles from a historical newspaper collection. This dataset can be standalone or made of documents returned by a search engine for a given query $\mathcal{Q}$. The dataset could be in raw format or have been pre-processed. We would suggest at least the two following pre-processing steps: first, we recommend to clean the dataset if it has been processed with OCR, either manually or semi-automatically, since the OCR quality will impact further steps \cite{DBLP:journals/csur/NguyenJCD21}. Secondly, we recommend to detect temporal expressions, as they are a good indicator of event mentions. Temporal expressions are either explicit (e.g. February 17, 1995) or implicit (e.g. yesterday, next month). One can use tools such as HeidelTime \citep{strotgen-gertz-2010-heideltime} or SUTime \citep{chang-manning-2012-sutime} to detect temporal expressions in text and resolve them to an absolute date format, simplifying their use in the TLS process. However, we must keep in mind that the detection of temporal expressions, especially implicit ones, is still a challenging task. Moreover, available tools such as these were mainly conceived for contemporary data, and thus may not work as properly on historical data.

The input dataset could be pre-processed further by applying NLP components such as Name Entity Recognition (NER), Topic Modelling (TM), Event Extraction (EE), Relation Extraction (RE), Keyword Extraction (KE), or Keyword Generation (KG). Such annotations could be used to index the dataset and allow the user to query documents about a specific Named Entity or topic, as in the impresso\footnote{\url{https://impresso-project.ch/app/}} or the NewsEye\footnote{\url{https://www.newseye.eu/}} platforms.

% ===========================================================

\vspace{-.5em}
\subsection{Timeline Generation}
\label{timelinegen}
\vspace{-.5em}

In this section, we present the first main step of the framework, which extracts mentions of events and attributes them an importance score. 
\vspace{-.5em}
\subsubsection{Event Detection}
\label{sec:eventdet}

Although events can be defined in many ways, a commonly accepted definition is "something that is \textit{happening} or that is holding true in a given circumstance", as stated in the TimeML guidelines \citet{timeml-guidelines}. Events can be detected in multiple ways: one could detect them through statistical analysis of the corpus. For instance, \citet{10.1145/1008992.1009065} measure the occurrences of similar sentences associated with the same date, whereas \citet{10.1007/978-3-030-15719-7_34} measure the occurrences of articles in atomic time intervals to later aggregate them and determine the bursty time periods. These statistical methods are especially suited for homogeneous datasets, but may not work as well on heterogeneous or fragmentary datasets. One could also train a Learning-to-Rank model on summaries created by experts in order to detect important sentences as in \citep{Tran2013LeveragingLT}. This would, however, require training data which tend to be scarce, even when for contemporary data.

Alternatively, one could use an Event Detection model to detect and annotate events in the dataset as in \citet{chasin-2010-event-wiki-linea}. Event Detection is another task in the NLP community that has been extensively studied, and some previous works such as \citet{DBLP:conf/aiia/NguyenBLD19} have already applied these methods in humanities contexts. However, we need to keep in mind that training such a model requires annotated resources that are often lacking, especially for historical data, and that the OCR quality of documents impacts the output of these models.

Finally, we could select as event any sentence containing at least a time expression, either explicit or implicit as in \citet{10.1007/978-3-030-26075-0_27, nguyen-etal-2014-ranking}. This selection could be made even finer by taking sentences that also contain a Named Entity as in \citet{abujabal-events-2015, bedi-etal-2017-event}. One can then apply algorithms such as Affinity Propagation \citep{doi:10.1126/science.1136800} or Chinese Whispers \citep{biemann-2006-chinese} to gather sentences describing the same event as in \citet{rusu-etal-2014-unsupervised, yu-etal-2021-multi, steen-markert-2019-abstractive}.

Regardless of the method used to detect them, events should all be associated with time. These could be the time expressions occurring with the event mentions, or the Document Creation Date (DCD) if no time expressions are present. 
%Moreover, we think the method to detect events could be selected from the user interface. 
%A default option could be any sentence containing a temporal expression, but if the dataset is for instance pre-processed with Event Detection techniques, the obtained annotations could be used instead. 
Alternatively, approaches for estimating the focus time of text \cite{DBLP:journals/ipm/JatowtYT15}, in absence of any temporal expressions can be applied to associate event-related sentences with particular points of time.

\subsubsection{Event Importance Estimation}
\label{sec:eventimp}

As mentioned in Section \ref{sec:relatedwork}, the importance of an event can be measured in a supervised or semi-supervised manner with a classifier \citep{chasin-2010-event-wiki-linea, ge-etal-2015-bring}. This method, however, requires training data that are difficult to obtain or produce. Furthermore, the process leading a classifier to a prediction is generally not explained. Since the goal of this framework is to assist in the study of longitudinal datasets, it is necessary that the process of generating a timeline is interpretable. Thus, we would suggest to measure the importance score in an unsupervised manner by extracting features from the dataset as in \citep{nguyen-etal-2014-ranking, 10.1145/1008992.1009065, 10.1007/978-3-319-76941-7_80}. Some of the features that we think could help measure this importance score are listed below, with suggestions on how to compute them:

\vspace{-.5em}
\begin{description}
                
        \item[Redundancy:] The more frequently an event is mentioned, the more important it should be. One can then simply count the occurrences of events, or as an alternative, assign them importance weights by calculating their TF-IDF scores over all the time units. However, as the data might be fragmentary in archive datasets, this feature should rather not be used alone;
        
    \vspace{-6pt}

        \item[Contemporary references:] an event may be important at a given time if other events occurring around the same period of time refer to it. Thus, to evaluate this feature, we could count how often an event is referred to from the descriptions of other events in a given short period of time around that event;
    \vspace{-6pt}

        \item[Retrospective references:] Similarly, an event is likely to be important if documents keep mentioning it some time after it occurred. To assess this kind of across-time reference to the event, one could count how often (and perhaps for how long) an event is mentioned by other events that occurred after a given period of time. Other solutions may rely on computing random walks over graphs composed of timestamped events and/or entities to measure the amount of signal propagation from the past towards "the recent times" \cite{DBLP:conf/cikm/JatowtKT16};
      \vspace{-12pt}

        \item[Causality:] an event is likely to be important if it is the cause of other events that occurred after it. To evaluate the causality of an event, one could use \textit{date reference graphs} as in \citet{tran-etal-2015-joint}, which measure the frequency of references, the topical influence and temporal influence between two events to determine a causal link. It is also possible to use Causal Relation Extraction (CRE) methods as presented by \citet{osti_10089455} for example. However, the CRE task is far from solved and may require much more dataset pre-processing;
        
            \vspace{-8pt}
            
        \item[Common sense:] some events are clearly more important than other, e.g. the birth of a child or marrying a partner are usually more important events in a family history than repainting a house. To represent that kind of common sense knowledge and compute this feature, it may be necessary to create a dataset of events that are deemed important to train a 1-class classifier (1CC) as in \citet{10.1007/978-3-030-26075-0_27} or a Learning-to-Rank model as in \citet{ge-etal-2015-bring}. Note that while important events can be collected from historical textbooks or history-related content, gathering unimportant events may be less easy and more problematic; hence the solution could be to rely on a 1CC task. 
         % \vspace{-6pt}

\end{description}

% Relevance different than importance

Using these features, a straightforward formula to calculate the importance of an event could be: \[\alpha \cdot F1 + \beta \cdot F2 + \gamma \cdot F3 + \delta \cdot F4 + \epsilon \cdot F5\] 
where $F1, F2, F3, F4, F5$ are the scaled values of the features described above and $\alpha, \beta, \gamma, \delta, \epsilon$ are hyper-parameters of which value is defined by the user or document archive custodians. Similarly to event detection, the user could be asked to select any of these features to compute this score. 

Some periods may contain much more documents than others. For instance, fewer documents may be available during a war time because of censorship or paper restriction. This lack of documents may lead to events that are far more or far less mentioned than others, and bias frequency-based features such as \textit{redundancy}, \textit{contemporary} and \textit{retrospective  references}. Thus, these features should be normalized before being incorporated.

Furthermore, we suggest these features since they are easy to compute, but we also acknowledge that they may not be sufficient to measure the importance of an event from the perspective of an expert such as a historian. Because the formula to compute the importance score is modular, one could incorporate more features in collaboration with experts.

\vspace{-.5em}
\subsection{Timeline Presentation}
\label{sec:timelinepresentation}
\vspace{-.5em}
In this section, we describe the second main step of the framework, which generates the timeline from events scored in the previous step. We present sets of filters to select which events should appear on the timeline and how they should be presented. We also describe an optional step of timeline augmentation using external data. 

\vspace{-.5em}
\subsubsection{Event Filtering}
\label{sec:eventfilter}

A dataset may contain hundreds or thousands of mentioned events. It is necessary to select those that will be added to the timeline. To do so, we can use filters such as described below. The weight of these filters could be changed on the user interface, thus allowing users to instantly update the timeline.

\begin{description}
    \item[Top $N$:] top $N$ most important events are retained;
    \vspace{-16pt}
    
    \item[Importance Threshold ($IT$):] only events of which the importance score is superior to a pre-fixed threshold $IT$ are taken. Individual thresholds for the features described in Section \ref{timelinegen} that make up the importance score can also be set;
    \vspace{-10pt}
    
    \item[Topical Diversity Threshold ($TopDT$):]  removes redundant event mentions and ensures the timeline is topically diverse. Topical diversity can be simply measured using Maximal Marginal Relevance (MMR) \citep{GoldsteinStewart1998SummarizationU} or the \textit{n}-gram blocking metric as in \citet{10.48550/arxiv.1903.10318};
    \vspace{-6pt}
    
    \item[Temporal Diversity Threshold ($TempDT$)]: ensures every time unit on the generated timeline is evenly represented by setting a minimum and maximum number of events that can appear at each time unit.
    
\end{description}
\vspace{-.5em}

\vspace{-.5em}
\subsubsection{Event Description Selection}
\label{sec:eventdesc}

There are multiple ways to represent an event on a timeline. 
%Similar to the filters, the user could instantly update the timeline on the interface by selecting a description. 
One could select a sentence that describes the event. If this sentence is too long, one could use sentence compression methods \citep{Filippova2008DependencyTB} to only keep its most important part. As mentioned earlier, an event might be represented by a cluster of sentences. The user can thus select one sentence among this cluster or generate a cloud of terms of all sentences contained in it, as in \citet{10.1007/978-3-030-26075-0_27}. One could also use headlines if the target documents are articles as in \citet{tran-alrifai-herder-headlines-2015, 10.1007/978-3-030-15719-7_34}.

Finally, we could also use a Natural Language Generation (NLG) system as in \citet{steen-markert-2019-abstractive}, as these generated texts are often easier to understand than text extracted from the documents. However, abstractive methods such as these may suffer from inaccuracies or hallucinations, i.e. generate information that is not present in the original documents. Thus, abstractive methods might generate improper event descriptions and lose the connection with the original documents. On the other hand, a common drawback of purely extractive methods is that selected sentences may require some context or at least post-processing for users to be able to properly understand them (e.g. pronouns may need to be resolved or we need to add definitions or descriptions of some entities or events).

\vspace{-.5em}
\subsubsection{Timeline Augmentation}
\label{sec:tlaug}
%\vspace{-.5em}

To properly understand them, some events may require contextual knowledge that is missing from the processed dataset. This can especially happen if the user is not a domain expert. Such contextual knowledge may be found in knowledge bases such as Wikidata or Wikipedia Year pages (see for example \cite{DBLP:conf/wsdm/TranCKN15}). Thus, timelines generated by an ATLS system could be augmented with contextual data provided by external knowledge bases as in \cite{10.1145/2600428.2609526}. These augmented timelines could help in explaining a dataset by summarizing it and providing the user with the necessary knowledge to understand it. Unfortunately, most resources created by experts are not in a machine-readable format \citep{gutehrle:hal-03468926}. Hence, this step may require more effort.

\vspace{-.5em}
\subsection{Timeline Evaluation}
\vspace{-.5em}
As mentioned earlier, the evaluation of a TLS system is a difficult task because of the lack of evaluation datasets and the inherent subjectivity of the task. In order to evaluate the output, we would suggest to manually assess the produced timelines, either by following some evaluation criteria as in \citet{duan-jatowt-katsumi-multitl-2017}, or by comparing them with resources created by experts such as timelines derived from history books as in \citet{bedi-etal-2017-event}. One could also use this framework to bootstrap an evaluation dataset specific to the given corpus, towards an automatic evaluation. 

% ========================================================

\vspace{-.5em}
\section{Discussion}
\label{sec:discussion}
\vspace{-.5em}
In this section, we describe two hypothetical use cases comparing the application of TLS and ATLS systems, and compare in Table \ref{tab:comparisontable} the types of datasets and timelines both methods can process. Finally, we discuss potential extensions of ATLS.

\begin{table*}[]
\centering
\small
    \begin{tabular}{l|cc}
    
        \textbf{\textbf{}}            & \multicolumn{1}{c|}{\textbf{\textbf{TLS}}} & \textbf{\textbf{ATLS}} \\ \hline
        
        \textbf{Covered period}    & \multicolumn{1}{c|}{Shorter}          & Longer        \\ 
        \hline
        
        \textbf{Input Data Size}              & \multicolumn{1}{c|}{Small / Medium} & Large       \\ 
        \hline
        
        \textbf{Documents type}       & \multicolumn{2}{c}{Timestamped documents (e.g. news articles)}           \\ \hline
        
        \textbf{Document Format}              & \multicolumn{1}{c|}{Usually born digital} & Often digitized       \\ 
        \hline
        
        \textbf{Data Integrity}         & \multicolumn{1}{c|}{Usually complete}       & Can be fragmentary \\ 
        \hline
        
        % \textbf{Annotations}         & \multicolumn{1}{c|}{No}             & May be provided    \\ 
        % \hline
        
        \textbf{Presence of noise}      & \multicolumn{1}{c|}{Less likely}          & Depends on OCR quality \\ \hline
        
        \textbf{Semantic evolution} & \multicolumn{1}{c|}{Less common}                  & Possible (esp. over long time)              \\ \hline
        
        \textbf{Need for query-based filtering} & \multicolumn{2}{c}{Optional (depends on data size and heterogeneity)}                      \\ 
        \hline

         \textbf{Need for contextualization}     & \multicolumn{1}{c|}{Less likely}             & More likely (esp. over long time)           \\ 
        \hline
        
        \textbf{Need for interpretable output}    & \multicolumn{2}{c}{Yes}   
        % & \multicolumn{1}{c|}{Less}             & More (esp. over long time)        

    \end{tabular}
    \caption{Comparison of 
    %dataset types and timelines 
    TLS and ATLS tasks 
    %can process
    }
    \label{tab:comparisontable}

\end{table*}

\vspace{-.5em}
\subsection{Use cases}
\label{sec:usecases}
\vspace{-.5em}
In the first hypothetical use case, a user has curated a homogeneous dataset of timestamped documents from Web archives. This dataset is made of news articles related to a story spanning over a year. It has been pre-processed to remove HTML tags and extract temporal expressions. To generate the timeline, the user applies the TLS method: important dates are first selected before generating a summary of events occurring at each date. The user can select the sentence mentioning the event, the headline of the article or apply abstractive methods to generate its description.

In the second hypothetical use case, a user has curated a heterogeneous corpus to study the economical life of a certain French region in the 20th century. This corpus is composed of periodicals, newspapers and magazines from different sources (parishes, libraries, etc.) published over a century and processed with OCR. This dataset has also been pre-processed: the documents have been first cleaned of OCR errors, then automatically annotated with Temporal Expression Extraction and Named Entity Recognition components. Furthermore, the dataset has been indexed so as to allow query-based searching. To generate the timeline, the user applies some of the ATLS approaches mentioned in this paper: events are first detected by clustering similar sentences that contain a temporal expression and a Named Entity. The importance of these events is then scored using the formula described in Section \ref{sec:eventimp}. The timeline is generated by setting high values to the topical and temporal diversity thresholds, and augmented with external data from Wikidata, so as to ensure a comprehensive and contextualized timeline. Similarly, the user can select from the user interface to use a cloud of terms or a sentence from the cluster to generate event descriptions. 
% This timeline can be generated from the documents returned by a query, or from the whole dataset.

\vspace{-.5em}
\subsection{Extensions of ATLS systems}
\vspace{-.5em}
Timelines are usually represented linearly, where each time unit is of the same size (usually a day or a year). However, the optimal granularity of temporal units might vary when generating a timeline over a long period of time. For example, when referring to a distant past, humans tend to often describe entire decades or years rather than discussing each day or month which is more common for the recent past. Furthermore, events mentioned in historical documents might not always be recorded with the same temporal precision (e.g., some events may have missing dates, the dates can be imprecise or difficult to be inferred). A possible solution would be to generate logarithmic timelines, where the granularity of the time unit changes over time, as suggested in \citet{10.1145/2063576.2063759}.

If the documents in the datasets are annotated with Named Entities, one could generate entity-based timelines. This could help understand the history of a specific entity such as a person or a location as in \citet{10.1007/978-3-030-26075-0_27}. This idea could be extended by generating aggregate timelines for multiple entities at the same time. These timelines could be agglomerative or contrastive and respectively show the similarities and differences between the history of multiple entities of the same type (e.g., cities in the same region or country, scientists of the same area). Similar to \citet{Duan2020ComparativeTS}, such comparative timelines would allow to study the history of entities of the same or similar type, e.g. Berlin vs. Paris or even entities of different types, e.g. Paris and the writer Victor Hugo.

\vspace{-.5em}
\section{Conclusion}
\label{sec:conclusion}
\vspace{-.5em}
TimeLine Summarization can be a useful tool for getting an overview of historical collections as well as it can serve as a novel information access means to news article archives.
In this position paper, we have presented an overview of existing TLS methods and described a conceptual framework for Archive TimeLine Summarization systems. 
%We have first made multiple suggestions to implement this framework and interact with it through user interfaces before discussing potential use cases and extensions. We believe these solutions can generate informative, readable and interpretable timelines.
% , thus assisting in the study of historical collections.

The implementation of the framework outlined in this paper will be the subject of our future work. We also intend to ask humanities scholars (historians, archivists, ...) to evaluate the quality of generated timelines and the effectiveness of our framework for the study of archive collections. 
% Moreover, we will experiment with the potential extensions of this framework such as entity-based timeline generation. 
% Finally, we plan to evaluate subsidiary aspects of this framework, such as the impact of OCR on the quality of the generated timelines.

% evqluqte features

\bibliography{biblio}

\begin{thebibliography}{49}
\expandafter\ifx\csname natexlab\endcsname\relax\def\natexlab#1{#1}\fi

\bibitem[{Abujabal and Berberich(2015)}]{abujabal-events-2015}
Abdalghani Abujabal and Klaus Berberich. 2015.
\newblock \href {https://doi.org/10.1145/2740908.2741692} {Important events in
  the past, present, and future}.
\newblock pages 1315--1320.

\bibitem[{Allan et~al.(1998)Allan, Papka, and Lavrenko}]{10.1145/290941.290954}
James Allan, Ron Papka, and Victor Lavrenko. 1998.
\newblock \href {https://doi.org/10.1145/290941.290954} {On-line new event
  detection and tracking}.
\newblock In \emph{Proceedings of the 21st Annual International ACM SIGIR
  Conference on Research and Development in Information Retrieval}, SIGIR '98,
  page 37–45, New York, NY, USA. Association for Computing Machinery.

\bibitem[{Alonso et~al.(2021)Alonso, Marchesin, Najork, and
  Silvello}]{DBLP:conf/desires/2021}
Omar Alonso, Stefano Marchesin, Marc Najork, and Gianmaria Silvello, editors.
  2021.
\newblock \href {http://ceur-ws.org/Vol-2950} {\emph{Proceedings of the Second
  International Conference on Design of Experimental Search {\&} Information
  REtrieval Systems, Padova, Italy, September 15-18, 2021}}, volume 2950 of
  \emph{{CEUR} Workshop Proceedings}. CEUR-WS.org.

\bibitem[{Bedi et~al.(2017)Bedi, Patil, Hingmire, and
  Palshikar}]{bedi-etal-2017-event}
Harsimran Bedi, Sangameshwar Patil, Swapnil Hingmire, and Girish Palshikar.
  2017.
\newblock \href {https://aclanthology.org/W17-5912} {Event timeline generation
  from history textbooks}.
\newblock In \emph{Proceedings of the 4th Workshop on Natural Language
  Processing Techniques for Educational Applications ({NLPTEA} 2017)}, pages
  69--77, Taipei, Taiwan. Asian Federation of Natural Language Processing.

\bibitem[{Biemann(2006)}]{biemann-2006-chinese}
Chris Biemann. 2006.
\newblock \href {https://aclanthology.org/W06-3812} {{C}hinese whispers - an
  efficient graph clustering algorithm and its application to natural language
  processing problems}.
\newblock In \emph{Proceedings of {T}ext{G}raphs: the First Workshop on Graph
  Based Methods for Natural Language Processing}, pages 73--80, New York City.
  Association for Computational Linguistics.

\bibitem[{Boros et~al.(2022)Boros, Nguyen, Lejeune, and
  Doucet}]{boros:hal-03635985}
Emanuela Boros, Nhu~Khoa Nguyen, Ga{\"e}l Lejeune, and Antoine Doucet. 2022.
\newblock \href {https://doi.org/10.1007/s00799-022-00325-2} {{Assessing the
  impact of OCR noise on multilingual event detection over digitised
  documents}}.
\newblock \emph{{International Journal on Digital Libraries}}.

\bibitem[{Campos et~al.(2018)Campos, Mangaravite, Pasquali, Jorge, Nunes, and
  Jatowt}]{10.1007/978-3-319-76941-7_80}
Ricardo Campos, V{\'i}tor Mangaravite, Arian Pasquali, Al{\'i}pio~M{\'a}rio
  Jorge, C{\'e}lia Nunes, and Adam Jatowt. 2018.
\newblock Yake! collection-independent automatic keyword extractor.
\newblock In \emph{Advances in Information Retrieval}, pages 806--810, Cham.
  Springer International Publishing.

\bibitem[{Ceroni et~al.(2014)Ceroni, Tran, Kanhabua, and
  Nieder\'{e}e}]{10.1145/2600428.2609526}
Andrea Ceroni, Nam~Khanh Tran, Nattiya Kanhabua, and Claudia Nieder\'{e}e.
  2014.
\newblock \href {https://doi.org/10.1145/2600428.2609526} {Bridging temporal
  context gaps using time-aware re-contextualization}.
\newblock In \emph{Proceedings of the 37th International ACM SIGIR Conference
  on Research \& Development in Information Retrieval}, SIGIR '14, page
  1127–1130, New York, NY, USA. Association for Computing Machinery.

\bibitem[{Chang and Manning(2012)}]{chang-manning-2012-sutime}
Angel~X. Chang and Christopher Manning. 2012.
\newblock \href
  {http://www.lrec-conf.org/proceedings/lrec2012/pdf/284_Paper.pdf} {{SUT}ime:
  A library for recognizing and normalizing time expressions}.
\newblock In \emph{Proceedings of the Eighth International Conference on
  Language Resources and Evaluation ({LREC}'12)}, pages 3735--3740, Istanbul,
  Turkey. European Language Resources Association (ELRA).

\bibitem[{Chasin(2010)}]{chasin-2010-event-wiki-linea}
Rachel Chasin. 2010.
\newblock Event and temporal information extraction towards timelines of
  wikipedia articles.

\bibitem[{Chieu and Lee(2004)}]{10.1145/1008992.1009065}
Hai~Leong Chieu and Yoong~Keok Lee. 2004.
\newblock \href {https://doi.org/10.1145/1008992.1009065} {Query based event
  extraction along a timeline}.
\newblock In \emph{Proceedings of the 27th Annual International ACM SIGIR
  Conference on Research and Development in Information Retrieval}, SIGIR '04,
  page 425–432, New York, NY, USA. Association for Computing Machinery.

\bibitem[{Duan et~al.(2017)Duan, Jatowt, and
  Tanaka}]{duan-jatowt-katsumi-multitl-2017}
Yijun Duan, Adam Jatowt, and Katsumi Tanaka. 2017.
\newblock \href {https://doi.org/10.1145/3078714.3078725} {Discovering typical
  histories of entities by multi-timeline summarization}.
\newblock In \emph{Proceedings of the 28th ACM Conference on Hypertext and
  Social Media}, HT '17, page 105–114, New York, NY, USA. Association for
  Computing Machinery.

\bibitem[{Duan et~al.(2019)Duan, Jatowt, and
  Tanaka}]{10.1007/978-3-030-26075-0_27}
Yijun Duan, Adam Jatowt, and Katsumi Tanaka. 2019.
\newblock History-driven entity categorization.
\newblock In \emph{Web and Big Data}, pages 349--364, Cham. Springer
  International Publishing.

\bibitem[{Duan et~al.(2020)Duan, Jatowt, and Yoshikawa}]{Duan2020ComparativeTS}
Yijun Duan, Adam Jatowt, and Masatoshi Yoshikawa. 2020.
\newblock Comparative timeline summarization via dynamic affinity-preserving
  random walk.
\newblock In \emph{ECAI}.

\bibitem[{Filippova and Strube(2008)}]{Filippova2008DependencyTB}
Katja Filippova and Michael Strube. 2008.
\newblock Dependency tree based sentence compression.
\newblock In \emph{INLG}.

\bibitem[{Frey and Dueck(2007)}]{doi:10.1126/science.1136800}
Brendan~J. Frey and Delbert Dueck. 2007.
\newblock \href {https://doi.org/10.1126/science.1136800} {Clustering by
  passing messages between data points}.
\newblock \emph{Science}, 315(5814):972--976.

\bibitem[{Gao et~al.()Gao, Choubey, and Huang}]{osti_10089455}
Lei Gao, Prafulla~Kumar Choubey, and Ruihong Huang.
\newblock \href {https://par.nsf.gov/biblio/10089455} {Modeling document-level
  causal structures for event causal relation identification}.
\newblock \emph{Proceedings of the 2019 Conference of the North American
  Chapter of the Association for Computational Linguistics: Human Language
  Technologies, Volume 1 (Long Papers)}.

\bibitem[{Ge et~al.(2015)Ge, Pei, Ji, Li, Chang, and Sui}]{ge-etal-2015-bring}
Tao Ge, Wenzhe Pei, Heng Ji, Sujian Li, Baobao Chang, and Zhifang Sui. 2015.
\newblock \href {https://doi.org/10.3115/v1/P15-1056} {Bring you to the past:
  Automatic generation of topically relevant event chronicles}.
\newblock In \emph{Proceedings of the 53rd Annual Meeting of the Association
  for Computational Linguistics and the 7th International Joint Conference on
  Natural Language Processing (Volume 1: Long Papers)}, pages 575--585,
  Beijing, China. Association for Computational Linguistics.

\bibitem[{Gholipour~Ghalandari and
  Ifrim(2020)}]{gholipour-ghalandari-ifrim-2020-examining}
Demian Gholipour~Ghalandari and Georgiana Ifrim. 2020.
\newblock \href {https://doi.org/10.18653/v1/2020.acl-main.122} {Examining the
  state-of-the-art in news timeline summarization}.
\newblock In \emph{Proceedings of the 58th Annual Meeting of the Association
  for Computational Linguistics}, pages 1322--1334, Online. Association for
  Computational Linguistics.

\bibitem[{Goldstein-Stewart and
  Carbonell(1998)}]{GoldsteinStewart1998SummarizationU}
Jade Goldstein-Stewart and Jaime~G. Carbonell. 1998.
\newblock Summarization: (1) using {MMR} for {D}iversity-{B}ased {R}eranking
  and (2) {E}valuating {S}ummaries.
\newblock In \emph{TIPSTER}.

\bibitem[{Gutehrl{\'e} et~al.(2021)Gutehrl{\'e}, Harlamov, Karimi, Wei,
  Jean-Caurant, and Pivovarova}]{gutehrle:hal-03468926}
Nicolas Gutehrl{\'e}, Oleg Harlamov, Farimah Karimi, Haoyu Wei, Axel
  Jean-Caurant, and Lidia Pivovarova. 2021.
\newblock \href {https://hal.archives-ouvertes.fr/hal-03468926} {{SpaceWars: A
  Web Interface for Exploring the Spatio-temporal Dimensions of WWI Newspaper
  Reporting}}.
\newblock \emph{{CEUR Workshop Proceedings}}.

\bibitem[{Jatowt and Au~Yeung(2011)}]{10.1145/2063576.2063759}
Adam Jatowt and Ching-man Au~Yeung. 2011.
\newblock \href {https://doi.org/10.1145/2063576.2063759} {Extracting
  collective expectations about the future from large text collections}.
\newblock In \emph{Proceedings of the 20th ACM International Conference on
  Information and Knowledge Management}, CIKM '11, page 1259–1264, New York,
  NY, USA. Association for Computing Machinery.

\bibitem[{Jatowt et~al.(2016)Jatowt, Kawai, and
  Tanaka}]{DBLP:conf/cikm/JatowtKT16}
Adam Jatowt, Daisuke Kawai, and Katsumi Tanaka. 2016.
\newblock Predicting importance of historical persons using wikipedia.
\newblock In \emph{Proceedings of the 25th {ACM} International Conference on
  Information and Knowledge Management, {CIKM} 2016, Indianapolis, IN, USA,
  October 24-28, 2016}, pages 1909--1912. {ACM}.

\bibitem[{Jatowt et~al.(2015)Jatowt, Yeung, and
  Tanaka}]{DBLP:journals/ipm/JatowtYT15}
Adam Jatowt, Ching{-}man~Au Yeung, and Katsumi Tanaka. 2015.
\newblock Generic method for detecting focus time of documents.
\newblock \emph{Inf. Process. Manag.}, 51(6):851--868.

\bibitem[{Kessler et~al.(2012)Kessler, Tannier, Hag{\`e}ge, Moriceau, and
  Bittar}]{Kessler2012FindingSD}
R{\'e}my Kessler, Xavier Tannier, Caroline Hag{\`e}ge, V{\'e}ronique Moriceau,
  and Andr{\'e} Bittar. 2012.
\newblock Finding salient dates for building thematic timelines.
\newblock In \emph{ACL}.

\bibitem[{La~Quatra et~al.(2021)La~Quatra, Cagliero, Baralis, Messina, and
  Montagnuolo}]{quatra-al-datefirst-2021}
Moreno La~Quatra, Luca Cagliero, Elena Baralis, Alberto Messina, and Maurizio
  Montagnuolo. 2021.
\newblock \href {https://doi.org/10.1145/3404835.3462954} {\emph{Summarize
  Dates First: A Paradigm Shift in Timeline Summarization}}, page 418–427.
  Association for Computing Machinery, New York, NY, USA.

\bibitem[{Lin(2004)}]{lin-2004-rouge}
Chin-Yew Lin. 2004.
\newblock \href {https://aclanthology.org/W04-1013} {{ROUGE}: A package for
  automatic evaluation of summaries}.
\newblock In \emph{Text Summarization Branches Out}, pages 74--81, Barcelona,
  Spain. Association for Computational Linguistics.

\bibitem[{Linhares~Pontes et~al.(2019)Linhares~Pontes, Hamdi, Sid{\`e}re, and
  Doucet}]{linharespontes:hal-02557116}
Elvys Linhares~Pontes, Ahmed Hamdi, Nicolas Sid{\`e}re, and Antoine Doucet.
  2019.
\newblock \href {https://doi.org/10.1007/978-3-030-34058-2\_11} {{Impact of OCR
  Quality on Named Entity Linking}}.
\newblock In \emph{{International Conference on Asia-Pacific Digital Libraries
  2019}}, Kuala Lumpur, Malaysia.

\bibitem[{Liu(2019)}]{10.48550/arxiv.1903.10318}
Yang Liu. 2019.
\newblock \href {https://doi.org/10.48550/ARXIV.1903.10318} {Fine-tune {BERT}
  for extractive summarization}.

\bibitem[{Martschat and Markert(2018)}]{martschat-markert-2018-temporally}
Sebastian Martschat and Katja Markert. 2018.
\newblock \href {https://doi.org/10.18653/v1/K18-1023} {A temporally sensitive
  submodularity framework for timeline summarization}.
\newblock In \emph{Proceedings of the 22nd Conference on Computational Natural
  Language Learning}, pages 230--240, Brussels, Belgium. Association for
  Computational Linguistics.

\bibitem[{Mihalcea and Tarau(2004)}]{mihalcea-tarau-2004-textrank}
Rada Mihalcea and Paul Tarau. 2004.
\newblock \href {https://aclanthology.org/W04-3252} {{T}ext{R}ank: Bringing
  order into text}.
\newblock In \emph{Proceedings of the 2004 Conference on Empirical Methods in
  Natural Language Processing}, pages 404--411, Barcelona, Spain. Association
  for Computational Linguistics.

\bibitem[{Minard et~al.(2015)Minard, Speranza, Agirre, Aldabe, van Erp,
  Magnini, Rigau, and Urizar}]{minard-etal-2015-semeval}
Anne-Lyse Minard, Manuela Speranza, Eneko Agirre, Itziar Aldabe, Marieke van
  Erp, Bernardo Magnini, German Rigau, and Rub{\'e}n Urizar. 2015.
\newblock \href {https://doi.org/10.18653/v1/S15-2132} {{S}em{E}val-2015 task
  4: {T}ime{L}ine: Cross-document event ordering}.
\newblock In \emph{Proceedings of the 9th International Workshop on Semantic
  Evaluation ({S}em{E}val 2015)}, pages 778--786, Denver, Colorado. Association
  for Computational Linguistics.

\bibitem[{Mutuvi et~al.(2018)Mutuvi, Doucet, Odeo, and
  Jatowt}]{mutuvi:hal-03025563}
Stephen Mutuvi, Antoine Doucet, Moses Odeo, and Adam Jatowt. 2018.
\newblock \href {https://doi.org/10.1007/978-3-030-04257-8\_1} {{Evaluating the
  Impact of OCR Errors on Topic Modeling}}.
\newblock In \emph{{Maturity and Innovation in Digital Libraries. 20th
  International Conference on Asia-Pacific Digital Libraries, ICADL 2018,
  Hamilton, New Zealand, November 19-22, 2018, Proceedings}}, pages 3 -- 14.

\bibitem[{Nguyen et~al.(2014)Nguyen, Tannier, and
  Moriceau}]{nguyen-etal-2014-ranking}
Kiem-Hieu Nguyen, Xavier Tannier, and Veronique Moriceau. 2014.
\newblock \href {https://aclanthology.org/C14-1114} {Ranking multidocument
  event descriptions for building thematic timelines}.
\newblock In \emph{Proceedings of {COLING} 2014, the 25th International
  Conference on Computational Linguistics: Technical Papers}, pages 1208--1217,
  Dublin, Ireland. Dublin City University and Association for Computational
  Linguistics.

\bibitem[{Nguyen et~al.(2020)Nguyen, Boros, Lejeune, and
  Doucet}]{DBLP:conf/aiia/NguyenBLD19}
Nhu~Khoa Nguyen, Emanuela Boros, Ga{\"{e}}l Lejeune, and Antoine Doucet. 2020.
\newblock \href {http://ceur-ws.org/Vol-2735/paper28.pdf} {Impact analysis of
  document digitization on event extraction}.
\newblock In \emph{Proceedings of the 4th Workshop on Natural Language for
  Artificial Intelligence {(NL4AI} 2020) co-located with the 19th International
  Conference of the Italian Association for Artificial Intelligence (AI*IA
  2020), Anywhere, November 25th-27th, 2020}, volume 2735 of \emph{{CEUR}
  Workshop Proceedings}, pages 17--28. CEUR-WS.org.

\bibitem[{Nguyen et~al.(2021)Nguyen, Jatowt, Coustaty, and
  Doucet}]{DBLP:journals/csur/NguyenJCD21}
Thi Tuyet~Hai Nguyen, Adam Jatowt, Micka{\"{e}}l Coustaty, and Antoine Doucet.
  2021.
\newblock Survey of {Post-OCR} processing approaches.
\newblock \emph{{ACM} Comput. Surv.}, 54(6):124:1--124:37.

\bibitem[{Pasquali et~al.(2021)Pasquali, Campos, Ribeiro, Santana, Jorge, and
  Jatowt}]{Pasquali2021TLSCovid19AN}
Arian Pasquali, Ricardo Campos, Alexandre Ribeiro, Brenda~Salenave Santana,
  Al{\'i}pio~M{\'a}rio Jorge, and Adam Jatowt. 2021.
\newblock Tls-covid19: A new annotated corpus for timeline summarization.
\newblock In \emph{ECIR}.

\bibitem[{Pasquali et~al.(2019)Pasquali, Mangaravite, Campos, Jorge, and
  Jatowt}]{10.1007/978-3-030-15719-7_34}
Arian Pasquali, V{\'i}tor Mangaravite, Ricardo Campos, Al{\'i}pio~M{\'a}rio
  Jorge, and Adam Jatowt. 2019.
\newblock Interactive system for automatically generating temporal narratives.
\newblock In \emph{Advances in Information Retrieval}, pages 251--255, Cham.
  Springer International Publishing.

\bibitem[{Rusu et~al.(2014)Rusu, Hodson, and
  Kimball}]{rusu-etal-2014-unsupervised}
Delia Rusu, James Hodson, and Anthony Kimball. 2014.
\newblock \href {https://doi.org/10.3115/v1/W14-2905} {Unsupervised techniques
  for extracting and clustering complex events in news}.
\newblock In \emph{Proceedings of the Second Workshop on {EVENTS}: Definition,
  Detection, Coreference, and Representation}, pages 26--34, Baltimore,
  Maryland, USA. Association for Computational Linguistics.

\bibitem[{Saurí et~al.(2006)Saurí, Moszkowicz, Knippen, Gaizauskas, Setzer,
  and Pustejovsky}]{timeml-guidelines}
Roser Saurí, Jessica Moszkowicz, Bob Knippen, Rob Gaizauskas, Andrea Setzer,
  and James Pustejovsky. 2006.
\newblock Timeml annotation guidelines version 1.2.1.

\bibitem[{Smith and Crane(2001)}]{10.5555/646634.699911}
David~A. Smith and Gregory Crane. 2001.
\newblock Disambiguating geographic names in a historical digital library.
\newblock In \emph{Proceedings of the 5th European Conference on Research and
  Advanced Technology for Digital Libraries}, ECDL '01, page 127–136, Berlin,
  Heidelberg. Springer-Verlag.

\bibitem[{Steen and Markert(2019)}]{steen-markert-2019-abstractive}
Julius Steen and Katja Markert. 2019.
\newblock \href {https://doi.org/10.18653/v1/D19-5403} {Abstractive timeline
  summarization}.
\newblock In \emph{Proceedings of the 2nd Workshop on New Frontiers in
  Summarization}, pages 21--31, Hong Kong, China. Association for Computational
  Linguistics.

\bibitem[{Str{\"o}tgen and Gertz(2010)}]{strotgen-gertz-2010-heideltime}
Jannik Str{\"o}tgen and Michael Gertz. 2010.
\newblock \href {https://aclanthology.org/S10-1071} {{H}eidel{T}ime: High
  quality rule-based extraction and normalization of temporal expressions}.
\newblock In \emph{Proceedings of the 5th International Workshop on Semantic
  Evaluation}, pages 321--324, Uppsala, Sweden. Association for Computational
  Linguistics.

\bibitem[{Swan and Allan(2000)}]{swan-allan-overwiew-2000}
Russell Swan and James Allan. 2000.
\newblock \href {https://doi.org/10.1145/345508.345546} {Automatic generation
  of overview timelines}.
\newblock In \emph{Proceedings of the 23rd Annual International ACM SIGIR
  Conference on Research and Development in Information Retrieval}, SIGIR '00,
  page 49–56, New York, NY, USA. Association for Computing Machinery.

\bibitem[{Tran et~al.(2015{\natexlab{a}})Tran, Alrifai, and
  Herder}]{tran-alrifai-herder-headlines-2015}
Giang Tran, Mohammad Alrifai, and Eelco Herder. 2015{\natexlab{a}}.
\newblock Timeline summarization from relevant headlines.
\newblock In \emph{Advances in Information Retrieval}, pages 245--256, Cham.
  Springer International Publishing.

\bibitem[{Tran et~al.(2015{\natexlab{b}})Tran, Herder, and
  Markert}]{tran-etal-2015-joint}
Giang Tran, Eelco Herder, and Katja Markert. 2015{\natexlab{b}}.
\newblock \href {https://doi.org/10.3115/v1/P15-1154} {Joint graphical models
  for date selection in timeline summarization}.
\newblock In \emph{Proceedings of the 53rd Annual Meeting of the Association
  for Computational Linguistics and the 7th International Joint Conference on
  Natural Language Processing (Volume 1: Long Papers)}, pages 1598--1607,
  Beijing, China. Association for Computational Linguistics.

\bibitem[{Tran et~al.(2013)Tran, Tran, Tran, Alrifai, and
  Kanhabua}]{Tran2013LeveragingLT}
Giang~Binh Tran, Tuan Tran, Nam~Khanh Tran, Mohammad Alrifai, and Nattiya
  Kanhabua. 2013.
\newblock Leveraging learning to rank in an optimization framework for timeline
  summarization.

\bibitem[{Tran et~al.(2015{\natexlab{c}})Tran, Ceroni, Kanhabua, and
  Nieder{\'{e}}e}]{DBLP:conf/wsdm/TranCKN15}
Nam~Khanh Tran, Andrea Ceroni, Nattiya Kanhabua, and Claudia Nieder{\'{e}}e.
  2015{\natexlab{c}}.
\newblock Back to the past: Supporting interpretations of forgotten stories by
  time-aware re-contextualization.
\newblock In \emph{Proceedings of the Eighth {ACM} International Conference on
  Web Search and Data Mining, {WSDM} 2015, Shanghai, China, February 2-6,
  2015}, pages 339--348. {ACM}.

\bibitem[{Yu et~al.(2021)Yu, Jatowt, Doucet, Sugiyama, and
  Yoshikawa}]{yu-etal-2021-multi}
Yi~Yu, Adam Jatowt, Antoine Doucet, Kazunari Sugiyama, and Masatoshi Yoshikawa.
  2021.
\newblock \href {https://doi.org/10.18653/v1/2021.acl-long.32}
  {Multi-{T}ime{L}ine summarization ({MTLS}): Improving timeline summarization
  by generating multiple summaries}.
\newblock In \emph{Proceedings of the 59th Annual Meeting of the Association
  for Computational Linguistics and the 11th International Joint Conference on
  Natural Language Processing (Volume 1: Long Papers)}, pages 377--387, Online.
  Association for Computational Linguistics.

\end{thebibliography}
\bibliographystyle{acl_natbib}

\end{document}